\documentclass[11pt,a4paper]{article}
\usepackage[hyperref]{acl2020}
\usepackage{times}
\usepackage{latexsym}
\usepackage{linguex} % for nice glossing
\usepackage{tabularx}
\usepackage{booktabs}
\usepackage{multirow,multicol}
\usepackage[utf8]{inputenc}
\usepackage{supertabular}
\usepackage{ stmaryrd }
\usepackage{subcaption}
\usepackage{amsmath}

\usepackage[]{url}
\usepackage{collcell}
\usepackage{hhline}
\usepackage{pgf}
\usepackage{float}

\usepackage{arydshln}
\usepackage{adjustbox}

\usepackage{graphicx} % to enable minipage for figs.
\graphicspath{{./images/}} % enables us to use subfolder named figs

\alignSubExtrue

\def\colorModel{hsb}
\newcommand\ColCell[1]{
  \pgfmathparse{#1<10?1:0}  %Threshold for changing the font color in the cells
    \ifnum\pgfmathresult=0\relax\color{white}\fi
    \pgfmathsetmacro\compA{135/360} %Component H (keep at 0 for b&w, 120 for green)
    \pgfmathsetmacro\compB{1}       %Component S (keep at 0 for b&w, 1 if using a single color)
    \pgfmathsetmacro\compC{(#1-25)/100}            %Component B (brightness - lower scores are more black)
  \edef\x{\noexpand\centering\noexpand\cellcolor[\colorModel]{\compA,\compB,\compC}}\x #1
  } 
\newcolumntype{E}{>{\collectcell\ColCell}m{4.5ex}<{\endcollectcell}}  %Cell width and defines column

\aclfinalcopy % Uncomment this line for the final submission
\newcommand{\DATASET}{\textsc{ImpPres}\xspace}
\def\newterm#1{\textbf{#1}}
 \usepackage{amssymb}% http://ctan.org/pkg/amssymb
 \usepackage{pifont}% http://ctan.org/pkg/pifont
 \newcommand{\cmark}{\ding{51}}%
 \newcommand{\xmark}{\ding{55}}%

\title{Are Natural Language Inference Models {\DATASET}sive? \\
Learning \textsc{Imp}licature and \textsc{Pres}upposition}% 

\author{
Paloma Jereti\v{c}\thanks{\ \ Equal Contribution}\hphantom{.}$^{1}$, Alex Warstadt\footnotemark[1]\hphantom{.}$^{1}$, Suvrat Bhooshan$^2$, Adina Williams$^2$\\
$^1$Department of Linguistics, New York University\\
$^2$Facebook AI Research\\
{\tt \{paloma,warstadt\}@nyu.edu, \{sbh,adinawilliams\}@fb.com}
}

\begin{document}
\maketitle
\begin{abstract}
%Natural language inference (NLI) is an increasingly important task for natural language understanding, which requires one to infer whether one sentence entails another. We assess whether systems trained under a common-sense NLI framework can generalize the notion of textual entailment to pragmatic inferences about a speaker's communicative intensions, which are needed for implicatures and presuppositions. We create an \textsc{Imp}licature and \textsc{Pres}upposition diagnostic dataset (\DATASET), consisting of 32,000 semi-automatically generated sentence pairs. Our resource can be seamlessly integrated into classic three-way NLI modeling efforts.
%We train three model (BOW, Infersent, and BERT) on NLI and test on \DATASET. 
%We find that BERT learns to make pragmatic inferences. BERT reliably treats implicatures triggered by ``some" and numerals as entailments. For some presupposition triggers such as \emph{only}, BERT reliably recognizes the presupposition as an entailment even when the trigger is embedded under an operator that cancels classical entailments, such as negation. We conclude that NLI training does encourage models to learn some pragmatic reasoning.

%We find that models fail to understand entailment-canceling operators (negation results accord with \citealt{mccoy2019right}, but we also show failures for modals, conditionals, interrogatives), and do generally poorly when embedded. 

Natural language inference (NLI) is an increasingly important task for natural language understanding, which requires one to infer whether a sentence entails another.
However, the ability of NLI models to make pragmatic inferences remains understudied. 
We create an \textsc{Imp}licature and \textsc{Pres}upposition diagnostic dataset (\DATASET), consisting of $>$25k semi-automatically generated sentence pairs illustrating well-studied pragmatic inference types. We use \DATASET to evaluate whether BERT, InferSent, and BOW NLI models trained on MultiNLI \cite{williams2018} learn to make pragmatic inferences. Although MultiNLI appears to contain very few pairs illustrating these inference types, we find that BERT learns to draw pragmatic inferences. It reliably treats scalar implicatures triggered by ``some" as entailments. For some presupposition triggers like \emph{only}, BERT reliably recognizes the presupposition as an entailment, even when the trigger is embedded under an entailment canceling operator like negation. BOW and InferSent show weaker evidence of pragmatic reasoning. We conclude that NLI training encourages models to learn some, but not all, pragmatic inferences.
\end{abstract}

\section{Introduction}

\begin{figure}[t]
    \centering
    \includegraphics[width=\columnwidth]{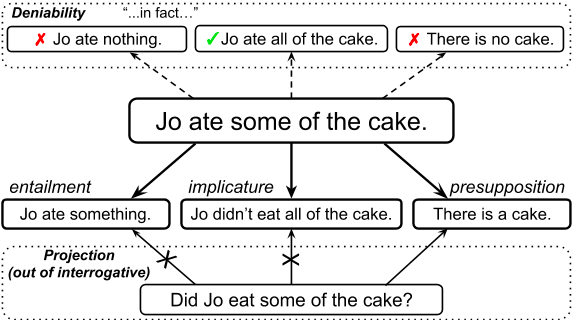}
    \caption{Illustration of key properties of classical entailments, implicatures, and presuppositions. Solid arrows indicate valid commonsense entailments, and arrows with X's indicate lack of entailment. Dashed arrows indicate follow up statements with the addition of \emph{in fact}, which can either be acceptable (marked with `{\color{red}\xmark}') or unacceptable (marked with `{\color{green}\cmark}').
}
    \label{fig:intro}
\end{figure}

%One of the most foundational semantic discoveries is that logical rules can be systematically found in the inferential relationships between pairs of natural language sentences \citep[\textit{De Interpretatione}, Ch. 6]{aristotle}. 
One of the most foundational semantic discoveries is that systematic rules govern the inferential relationships between pairs of natural language sentences \citep[\textit{De Interpretatione}, Ch. 6]{aristotle}. In natural language processing, Natural Language Inference (NLI)---a task whereby a system determines whether a pair of sentences instantiates in an entailment, a contradiction, or a neutral relation---has been useful for training and evaluating models on sentential reasoning.  However, linguists and philosophers now recognize that there are separate semantic and pragmatic modes of reasoning \citep{grice1975logic, clark1996, beaver1997presupposition, horn2004handbook,  potts2015presupposition}, and it is not clear which of these modes, if either, NLI models learn. We investigate two pragmatic inference types that are known to differ from classical entailment: scalar implicatures and presuppositions. As shown in Figure~\ref{fig:intro}, implicatures differ from entailments in that they can be denied, and presuppositions differ from entailments in that they are not canceled when placed in entailment-cancelling environments (e.g., negation, questions). 

To enable research into the relationship between NLI and pragmatic reasoning, we introduce \DATASET, a fine-grained NLI-style diagnostic test dataset for probing how well NLI models perform implicature and presupposition. Containing 25.5K sentence pairs illustrating key properties of these pragmatic inference types, \DATASET is automatically generated according to linguist-crafted templates, allowing us to create a large, lexically varied, and well controlled dataset targeting specific instances of both types.

%for which one sentence \defn{presupposes} the other, or for which one \defn{implicates} the negation of the other.
% Investigating these two, well-described sentential inference types in the context of NLI is important, because it enables targeted error analysis that can elucidate what semantic and pragmatic information about sentence relations was (or wasn't) learned by NLI models. While several of the largest recent NLI datasets likely contain a few, isolated examples of presuppositions and implicatures \citep{bowman2015,williams2018}\footnote{
% \citeauthor{williams2018} report that 22\% of their sentence pairs contain lexical triggers (such as \textit{regret} or \textit{stopped}) in the premise and/or hypothesis.}, no resource for sentence pair inference has been specifically annotated or constructed in such a way to enable detailed error analysis by inference type, making \DATASET the first such contribution of its kind. 

% To test whether NLI models learn to do well on \DATASET, 
We first investigate whether presuppositions and implicatures are present in NLI models' training data. We take MultiNLI \citep{williams2018} as a case study, and find it has few instances of pragmatic inference, and almost none that arise from specific lexical triggers (see \S \ref{subsec:praginMultiNLI}). Given this, we ask whether training on MultiNLI is sufficient for models to \emph{generalize} about these largely absent commonsense reasoning types. %  We use \DATASET to evaluate models trained on MultiNLI: (BOW, InferSent, BERT). 
We find that generalization is possible: the BERT NLI model shows evidence of pragmatic reasoning %on a high proportion of pragmatic reasoning 
when tested on the implicature from {\it some} to {\it not all}, and the presuppositions of certain triggers ({\it only}, cleft existence, possessive existence, questions). %Our negative results simply appear to be due to a  coarse-grained semantic knowledge of the relevant items.
We obtain some negative results, that suggest that models like BERT still lack a sophisticated enough understanding of the meanings of the lexical triggers for implicature and presupposition (e.g.,\ BERT  treats several word pairs as synonyms, e.g.,\ most notably, {\it or} and {\it and}).

%Our results show that BERT trained on MultiNLI largely passes basic controls probing its understanding of negation and basic entailment inferences, in contrast with simpler models such as BOW and InferSent that do poorly.
%Results for scalar implicatures overall reveal a yet coarse-grained semantic knowledge of the relationship between pairs of connectives, scalar adjectives, verbs, and modals, making us unable to conclude whether BERT behaves pragmatically.  However, 
%BERT produces informative results when tested on the determiner dataset (containing implicatures triggered by {\it some} and {\it not all}) that reveal a high proportion of pragmatic reasoning.
%Results for presuppositions BERT, and to some extent BOW, show a very high accuracy for certain triggers ({\it only}, cleft existence, possessive existence, questions), but not others (e.g. numerical triggers).
%These positive results show BERT's  ability to generalize from basic non-pragmatic inference types present in the data to well-behaved pragmatic inference, for some presuppositions and implicatures.

Our contributions are: (i) we provide a new diagnostic test set to probe for pragmatic inferences, complete with linguistic controls, (ii) to our knowledge, we present the first work evaluating deep NLI models on specific pragmatic inferences, and (iii) we show that BERT models can perform some types of pragmatic reasoning very well, even when trained on NLI data containing very few explicit examples of pragmatic reasoning. We publicly release all \DATASET data, models evaluated, annotations of MultiNLI, and the scripts used to process data.\footnote{\href{https://github.com/fairinternal/Imppres}{github.com/facebookresearch/ImpPres}}

\section{Background: Pragmatic Inference}\label{sec:background}

We take pragmatic inference to be a relation between two sentences relying on the utterance context and the conversational goals of interlocutors. Pragmatic inference contrasts with semantic entailment, which instead captures the logical relationship between isolated sentence meanings \citep{grice1975logic,stalnaker1974}.
We present \newterm{implicature} and \newterm{presupposition} inferences below.

\subsection{Implicature}\label{sec:implicature_background}

Broadly speaking, implicatures contrast with entailments in that they are inferences suggested by the speaker's utterance, but not included in its literal \citep{grice1975logic}. Although there are many types of implicatures we focus here on \newterm{scalar implicatures}. Scalar implicatures are inferences, often optional,\footnote{Implicature computation can depend on the cooperativity of the speakers, or on any aspect of the context of utterance (lexical, syntactic, semantic/pragmatic, discourse). See \citet{degen2015investigating} for a study of the high  variability of implicature computation, and the factors responsible for it.} which can be drawn when one member of a memorized lexical scale (e.g., $\langle$\textit{some, all}$\rangle$) is uttered (see \S \ref{SIdataset}). For example, when someone utters \emph{Jo ate some of the cake}, they suggest that \emph{Jo didn't eat all of the cake}, (see Figure~\ref{fig:intro} for more examples).
%TODO FIX "FOR MORE EXAMPLES"
According to Neo-Gricean pragmatic theory \cite{horn1989,levinson2000}, the inference \emph{Jo didn't eat all of the cake} arises because \emph{some} has a more informative lexical alternative \emph{all} that could have been uttered instead. We expect the speaker to make the most informative true statement:\footnote{This follows if we assume that speakers are cooperative \cite{grice1975logic} and knowledgeable \cite{gazdar1979pragmatics}.} as a result, the listener should infer that a stronger statement, where \emph{some} is replaced by \emph{all}, is false.%, or equivalently, the statement with \emph{not all} is true.

% belongs \newterm{lexical scales} \citep{horn1989,levinson2000}. Lexical scales are tuples of words ordered by informativity, i.e. where the second member of the scale asymmetrically entails (i.e. is stronger, or more informative than) the first, e.g. $\langle${\it some, all}$\rangle$, $\langle${\it or, and}$\rangle$, $\langle${\it can, have to}$\rangle$. 

% We illustrate the phenomenon with a scalar implicature arising from the use of {\it some}. 

% \vspace{-1ex}%

% \ex.\a.\label{impl.a} Jo ate {\bf some} of the cake.
% \
% \b.\label{impl.b} $\rightsquigarrow$ {\small \it Jo did \textbf{not} eat \textbf{all} of the cake.} \vspace{-1ex}%

% The utterance \ref{impl.a} suggests or {\it implicates} the meaning expressed by \ref{impl.b} (shown by the squiggly arrow), but does not express it literally. Scalar implicatures arise from the alternative utterances the speaker could have made: if the utterance contains a word that forms a scale with another word, an alternative utterance is built by substituting that word  with its scalemate. 

% The speaker is then assumed to pick the utterance that is maximally informative  \citep{grice1975logic}. The assertion `Jo ate some of the cake' is less informative than the alternative `Jo ate all of the cake'. If the alternative has been true, the speaker would have said it. They didn't, so the hearer can infer that the speaker takes it to be false. Thus, the implicature that Jo didn't eat all of the cake is computed.

Implicatures differ from entailments (and, as we will see, presuppositions; see Figure~\ref{fig:intro}) in that they are \newterm{deniable}, i.e.,\ they can be explicitly negated without resulting in a contradiction. For example, someone can utter \emph{Jo ate some of the cake}, followed by \emph{In fact, Jo ate all of it}. In this case, the implicature (i.e., \textit{Jo didn't eat all the cake} from above) has been denied. We thus distinguish implicated meaning from literal, or logical, meaning.

% We add a note on the broader notion of implicature, that is relevant to understanding the type of data present in the MultiNLI training dataset. 
% Scalar implicatures are a subset of conventional implicatures, i.e implicatures that are triggered by the meaning of specific words. These appear to be rare in the training data (see \S \ref{subsec:praginMultiNLI}). Far more common are \newterm{conversational implicatures}, that fully depend on the conversational context, and not on lexical meanings. We give an example in \ref{cvs}. 

% \ex.\label{cvs}
% \a.\label{cvs-assert} -- Will you have a cookie?
% \b.\label{cvs-impl} -- I am on a diet.\hfill $\rightsquigarrow$ {\small \it I will not have any.}

% The utterance in  \ref{cvs-impl} clearly suggests a negative answer to \ref{cvs-assert}, while not expressing it literally. It is a conversational implicature in that the inference does not depend on the specific words used, but the interaction of the discourse goals and the general content of the sentence. %Again, the inference arises from the assumption that speakers are cooperative, and utterances are relevant to the goals of the discourse.

\subsection{Presupposition}\label{sec:prsp}

\begin{table}[]
    \centering
    {\small
    \begin{tabular}{ll}
    \toprule
    Type & Example\\\midrule
        Trigger & Jo's cat yawned. \\
        Presupposition & Jo has a cat.\\\midrule
        Negated Trigger & Jo's cat didn't yawn.\\
        Modal Trigger & It's possible that Jo's cat yawned. \\
        Interrog. Trigger & Did Jo's cat yawn? \\
        Cond. Trigger & If Jo's cat yawned, it's OK. \\\midrule
        Negated Prsp. & Jo doesn't have a cat.\\
        Neutral Prsp. & Amy has a cat.\\\bottomrule
    \end{tabular}
    }
    \caption{Sample generated presupposition paradigm. Examples adapted from the `change-of-state' dataset.}
    \label{tab:prsp examples}
\end{table}

Presuppositions of a sentence are facts that the speaker takes for granted when uttering a sentence \citep{stalnaker1974,beaver1997presupposition}. Presuppositions are generally associated with the presence of certain expressions, known as \newterm{presupposition triggers}. For example, in Figure~\ref{fig:intro}, the definite description \emph{the cake} triggers the presupposition that there is a cake \citep{russell1905denoting}. Other examples of presupposition triggers are shown in Table~\ref{tab:prsp examples}.

Presuppositions differ from other inference types in that they generally \newterm{project} out of operators like questions and negation, meaning that they remain valid inferences even when \newterm{embedded under} these operators \citep{karttunen1973compound}.  The inference that there is a cake survives even when the presupposition trigger is in a question (\emph{Did Jordan eat some of the cake?}), as shown in Figure~\ref{fig:intro}. However, in questions, classical entailments and implicatures disappear. Table~\ref{tab:prsp examples} provides examples of triggers projecting out of several \newterm{entailment canceling operators}: negation, modals, interrogatives, and conditionals. 

It is necessary to clarify in what sense presupposition is a pragmatic inference. 
There is no consensus on whether presuppositions should  be considered part of the semantic content of expressions \citep[see][for opposing views]{stalnaker1974,heim1983projection}. However, presuppositions may come to be inferred via \newterm{accommodation}, a pragmatic process by which a listener infers the truth of some new fact based on its being presupposed by the speaker \cite{lewis1979scorekeeping}. For instance, if Jordan tells Harper that \textit{the King of Sweden wears glasses}, and Harper did not previously know that Sweden has a king, they would learn this fact by accommodation. With respect to NLI, any presupposition in the premise (short of world knowledge) will be new information, and therefore accommodation is necessary to recognize it as entailed. 

\section{Related Work}\label{subsec:RelatedWork}

NLI has been framed as a commonsense reasoning task \citep{dagan2006, Manning06}. One early formulation of NLI defines ``entailment'' as holding for sentences \textit{p} and \textit{h} whenever, ``typically, a human reading \textit{p} would infer that \textit{h} is most likely true\textellipsis[given] common human understanding of language [and] common background knowledge'' \citep{dagan2006}. Although this sparked debate regarding the terms \textit{inference} and \textit{entailment}---and whether an adequate notion of ``inference'' could be defined \citep{ZaenenKarttunenCrouch05, Manning06, CrouchKarttunenZaenen06}---in recent work, a commonsense formulation of ``inference'' is widely adopted \citep{bowman2015, williams2018} largely because it facilitates untrained annotators' participation in dataset creation.

NLI itself has been steadily gaining in popularity;  many  datasets for training and/or testing systems are now available including: FraCaS \citep{cooper1994}, RTE \citep{dagan2006, mirkin2009, dagan2013}, Sentences Involving Compositional Knowledge \citep[SICK]{marelli2014}, large scale imaging captioning as NLI \citep[SNLI]{bowman2015}, recasting other datasets into NLI \citep{glickman2006, white2017, poliak2018b}, ordinal commonsense inference \citep[JOCI]{zhang2017}, Multi-Premise Entailment \citep[MPE]{lai2017}, NLI over multiple genres of written and spoken English \citep[MultiNLI]{williams2018}, adversarially filtered common sense reasoning sentences \citep[(Hella)SWAG]{zellers2018, zellers2019}, explainable annotations for SNLI \citep[e-SNLI]{camburu2018}, cross-lingual NLI \citep[XNLI]{conneau2018}, scientific questioning answering as NLI \citep[SciTail]{khot2018}, NLI recast-question answering (part of \citealt[GLUE]{wang2019glue}), NLI for dialog \cite{welleck2019}, and NLI over narratives that require drawing inferences to the most plausible explanation from text \citep[$\alpha$NLI]{bhagavatula2020}. 
% \begin{figure}[t]
%     \centering
%     \includegraphics[width=\columnwidth]{InferenceUniverse}
%     \caption{Inference Datasets (mostly English).}
%     \label{fig:universe}
% \end{figure}
Other NLI datasets are created to identify where models fail \citep{glockner2018, naik-EtAl:2018:C18-1, mccoy2019right, schmitt2019}, many of which are also automatically generated \citep{geiger2018,yanaka2019b,yanaka2019,kim2019,nie2019, richardson2020}. %, from datasets that stymie NLI systems with lexical hypernym/hyponym relations

As datasets for NLI become increasingly numerous, one might wonder, do we need yet another NLI dataset? In this case, the answer is clearly yes: despite NLI's formulation as a commonsense reasoning task, it is still unknown whether this framing has resulted in models that learn specific modes of pragmatic reasoning. \DATASET is the first NLI dataset to explicitly probe whether models trained on commonsense reasoning actually do treat pragmatic inferences like implicatures and presuppositions as entailments without additional training on these specific inference types.

Beyond NLI, several recent works introduce resources for evaluating sentence understanding models for knowledge of pragmatic inferences. On the presupposition side, datasets such as MegaVeridicality \citep{white2018role} and CommitmentBank \citep{deMarneffe2019commitmentbank} compile gradient crowdsourced judgments regarding how likely a clause embedding predicate is to trigger a presupposition that its complement clause is true. \citet{white2018lexicosyntactic} and \citet{jiang2019you} find that LSTMs trained on a gradient event factuality prediction task on these respective datasets make systematic errors. Turning to implicatures, \newcite{degen2015investigating} introduces a dataset measuring the strength of the implicature from \emph{some} to \emph{not all} with crowd-sourced judgments. \newcite{schuster2019} find that an LSTM with supervision on this dataset can predict human judgments well. These resources all differ from \DATASET in two respects: First, their empirical scopes are all somewhat narrower, as all these datasets focus on only a single class of presupposition or implicature triggers. Second, the use of gradient judgments makes it non-trivial to use these datasets to evaluate NLI models, which are trained to make categorical predictions about entailment. Both approaches have advantages, and we leave a direct comparison for future work.

Outside the topic of sentential inference, \newcite{rashkin2018} propose a new task where a model must label actor intents and reactions for particular actions described using text. \citet{cianflone2018} create sentence-level adverbial presupposition datasets and train a binary classifier to detect contexts in which presupposition triggers (e.g., \textit{too}, \textit{again}) can be used.

\begin{table*}[ht!]
  \centering
{\small
\begin{tabular}{llllll}
\toprule
%\multicolumn{2}{l}{\bf Implicatures} & {\bf Relation Type}& \multicolumn{2}{c}{\bf Labels} & {\bf Item Type} \\
\bf Premise & \bf Hypothesis & \bf Relation type & \bf Logical label & \bf Pragmatic label & \bf Item type \\\midrule
\it some	& \it not all	& implicature ($+$ to $-$)
&\footnotesize \sf neutral	&\footnotesize \sf entailment&	target\\
\it not all& \it some	&implicature  ($-$ to $+$)
&\footnotesize \sf neutral		&\footnotesize \sf entailment&target\\
\midrule
\it some&\it 	all&	negated implicature 
($+$)
& \footnotesize \sf neutral	 &	\footnotesize \sf contradiction	&	target\\
\it all&\it 	some&	reverse negated implicature ($+$)
&	\footnotesize \sf entailment&	\footnotesize \sf contradiction&	target\\
\it not all&\it 	none&	negated implicature ($-$)
&	\footnotesize \sf neutral	 &	\footnotesize \sf contradiction	&target\\
\it none&\it 	not all&	reverse negated implicature ($-$)
&	\footnotesize \sf entailment &	\footnotesize \sf contradiction		&target\\ \midrule
\it all&\it 	none&	opposite	&\footnotesize \sf contradiction	&\footnotesize \sf contradiction	&	control\\
\it none&\it 	all&	opposite	&\footnotesize \sf contradiction	&\footnotesize \sf contradiction	&	control\\
\it some&	\it none&	 negation	&\footnotesize \sf contradiction	&\footnotesize \sf contradiction &		control\\
\it none&	\it some&negation	&\footnotesize \sf contradiction	&\footnotesize \sf contradiction	&	control\\
\it all&\it 	not all&	negation	&\footnotesize \sf contradiction	&\footnotesize \sf contradiction	&	control\\
\it not all&\it 	all&	 negation	&\footnotesize \sf contradiction	&\footnotesize \sf contradiction &		control\\
\bottomrule
\end{tabular}}
\caption{Paradigm for the scalar implicature datasets, with $\langle${\it some, all}$\rangle$ as an example.}\label{tab:paradigm_SI}
\end{table*}

\section{Annotating MultiNLI for Pragmatics}\label{subsec:praginMultiNLI}

In this section, we present results of an annotation effort that show that MultiNLI contains very little explicit evidence of pragmatic inferences of the type tested by \DATASET.
Although \citet{williams2018}\ report that 22\% of the MultiNLI development set sentence pairs contain lexical triggers (such as \textit{regret} or \textit{stopped}) in the premise and/or hypothesis, the mere presence of presupposition-triggering lexical items in the data does not show that MultiNLI contains evidence that presuppositions are entailments, since the sentential inference may focus on other types of information. To address this, 
we randomly selected 200 sentence pairs from the MultiNLI matched development set and presented them to three expert annotators with a combined total of 17 years of training in formal semantics and pragmatics.\footnote{{The full annotations are on the \DATASET repository.}} % Adina: 7 years %Paloma: 5 years 
Annotators answered the following questions for each pair: (1) are the sentences $P$ and $H$ related by a presupposition/implicature relation (entails/is entailed by, negated or not); (2) what subtype of inference (e.g.,\ existence presupposition, $\langle${\it some, all}$\rangle$ implicature); (3) is the presupposition trigger embedded under an entailment-cancelling operator?

Agreement among annotators was low, suggesting that few MultiNLI pairs are paradigmatic cases of implicatures or presuppositions. We found only 8 presupposition pairs and 3 implicature pairs on which two or more annotators agreed.  Moreover, we found only one example illustrating a particular inference type tested in \DATASET (the presupposition of possessed definites). All others were tagged as existence presuppositions and conversational implicatures (i.e. loose inferences dependent on world knowledge). %(10) (other types included: implicature from counterfactual, projected content of appositives, relevance implicatures, etc.). One presupposition was under an entailment-cancelling operator. 
The union of annotations was much larger: 42\% of examples were identified by at least one annotator as a presupposition or implicature (51 presuppositions and 42 implicatures, with 10 sentences receiving divergent tags). However, of these, only 23 presuppositions and 19 implicatures could reliably be used to learn pragmatic inference
(in 14 cases, the given tag did not match the pragmatic inference, and in 27 cases, computing the inference did not affect the relation type). Again, the large majority of implicatures were conversational, and most presuppositions were existential, and generally not linked to particular lexical triggers (e.g., topic marking).

We conclude that the MultiNLI dataset at best contains some evidence of loose pragmatic reasoning based on world knowledge and discourse structure, but almost no explicit information relevant to lexically triggered pragmatic inference, which is of the type tested in this paper.

\section{Methods}

\paragraph{Data Generation.} \DATASET consists of semi-automatically generated pairs of sentences with NLI labels illustrating key properties of implicatures and presuppositions.  We generate \DATASET using a codebase developed by \citet{warstadt2019npi} and significantly expanded for the BLiMP dataset \citep{warstadt2019blimp}. The codebase, including our scripts and documentation, are publicly available.\footnote{\href{https://github.com/alexwarstadt/data_generation}{github.com/alexwarstadt/data\_generation}}
Each sentence type in \DATASET is generated according to a template that specifies the linear order of the constituents in the sentence. The constituents are sampled from a vocabulary of over 3000 lexical items annotated with grammatical features needed to ensure morphological, syntactic, and semantic well-formedness. All sentences generated from a given template are structurally analogous up to the specified constituents, but may vary in sub-constituents. For instance, if the template calls for a verb phrase, the generated constituent may include a direct object or complement clause, depending on the argument structure of the sampled verb. See \S \ref{SIdataset} and \ref{subsec:presuppdata} for descriptions of the sentence types in the implicature and presupposition data.

Generating data lets us control the lexical and syntactic content so that we can guarantee that the sentence pairs in \DATASET evaluate the desired phenomenon \citep[see][for related discussion]{ettinger2016probing}. Furthermore, the codebase we use allows for greater lexical and syntactic variety than in many other templatic datasets \citep[see discussion in][]{warstadt2019blimp}. One limitation of this methodology is that generated sentences, while generally grammatical, often describe highly unlikely scenarios, or include low frequency combinations of lexical items (e.g., \emph{Sabrina only reveals this pasta}). Another limitation is that generated data is of limited use for training models, since it contains simple regularities that supervised classifiers may learn to exploit. Thus, we create \DATASET solely for the purpose of evaluating NLI models trained on standard datasets like MultiNLI.

% The same codebase is used for both the implicature and presupposition parts of \DATASET (

\paragraph{Models.} Our experiments evaluate NLI models trained on MultiNLI and built on top of three sentence encoding models: a bag of words (BOW) model, InferSent \citep{conneau2017}, and BERT-Large \citep{devlin2019bert}. The BOW and InferSent models use 300D GloVe embeddings as word representations \citep{pennington2014glove}. For the BOW baseline, word embeddings for premise and hypothesis are separately summed to create sentence representations, which are concatenated to form a single sentence-pair representation which is fed to a logistic regression softmax classifier. For the InferSent model, GloVe embeddings for the words in premise and hypothesis are respectively fed into a bidirectional LSTM, after which we concatenate the representations for premise and hypothesis, their difference, and their element-wise product \citep{mou2015}.\ BERT is a multilayer bidirectional transformer pretrained with the masked language modelling and next sequence prediction objectives, and finetuned on the MultiNLI dataset. We concatenate the premise and hypothesis after a special [\textsc{CLS}] token and separated them with the [\textsc{SEP}] token. The BERT representation for the [\textsc{CLS}] token is fed into classifier. We use Huggingface's pre-trained BERT trained on Toronto books \citep{zhu2015}.\footnote{\href{https://github.com/huggingface/pytorch-pretrained-BERT/}{github.com/huggingface/pytorch-pretrained-BERT/}} 

The BOW and InferSent models have development set accuracies of 49.6\% %(comparable to test set accuracy of 55.7\% from \citealt{conneau2017}) 
and 67.6\%. The development set accuracy for BERT-Large on MultiNLI is 86.6\%, similar to the results achieved by \citep{devlin2019bert}, but somewhat lower than state-of-the-art (currently 90.8\% on test from the ensembled RoBERTa model with long pretraining optimization, \citealt{liu2019}).

\section{Experiment 1: Scalar Implicatures}\label{sec:exp1}

% \paragraph{Implicature Data}

%We are not intending for this to be a balanced test set that is good for training models, but instead more of a benchmark test-set that enables model evaluation.  
% Testing both presuppositions and implicatures requires sets of negated sentences, to determine whether models perform correctly on basic semantic entailment controls given the lexical content of the sentences. Once that has been determined, we can subsequently investigate model behavior for the more complicated pragmatic inferences. The procedure for creating test and control sentence pairs for both implicatures and presuppositions proceeds by recombining types from templates. 
% Our test sets are larger and more expansive than the simple constituent negation test set in \citealt{kim2019} in that our negative operators interact with  entailment-canceling operators. \DATASET also as complementary to a recent negative polarity item (NPI) dataset \citep{warstadt2019npi}, which aimed to determine if popular models are sensitive to NPI licensing conditions, because it  accomplishes a different purpose. 

\subsection{Scalar Implicature Datasets}\label{SIdataset}

\begin{figure}[t]
    \centering
    \includegraphics[width=\columnwidth]{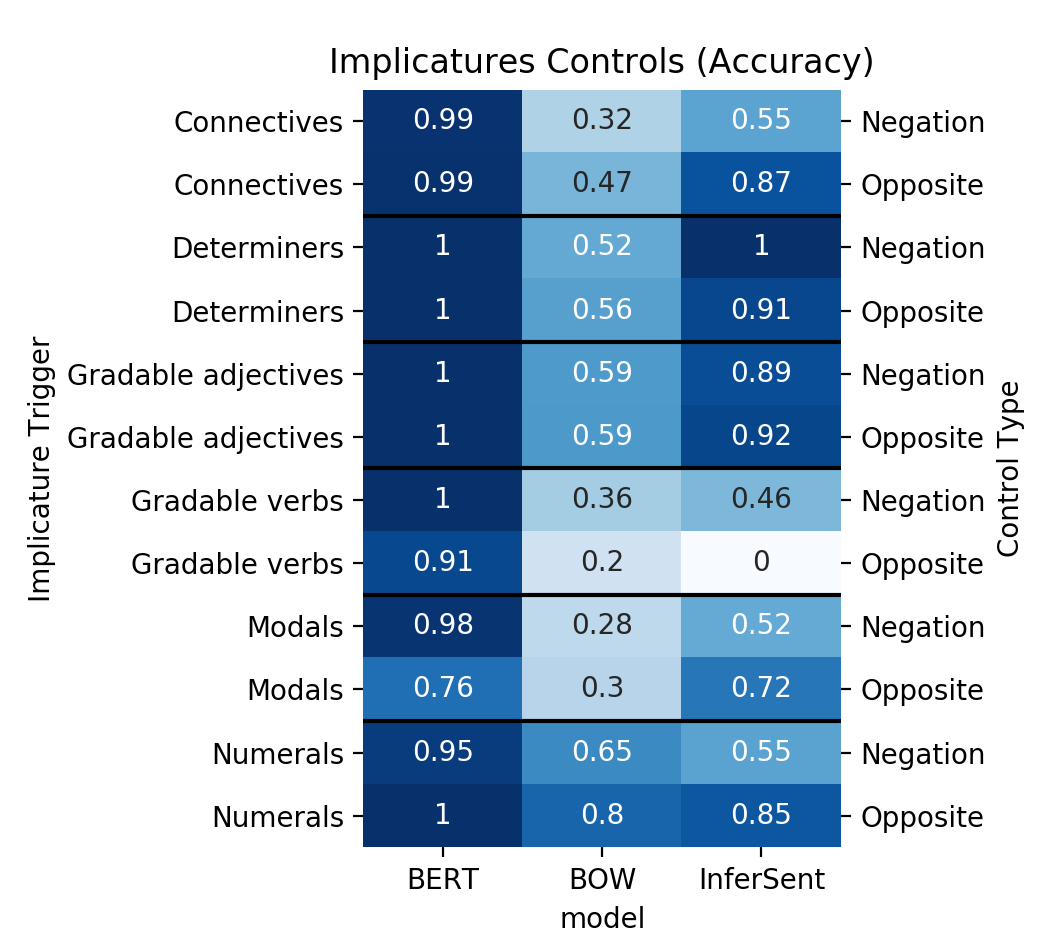}
    \caption{Results on Controls (Implicatures)}\label{fig:si_controls}
\end{figure}

\begin{figure}[t]
    \centering
    \includegraphics[width=\columnwidth]{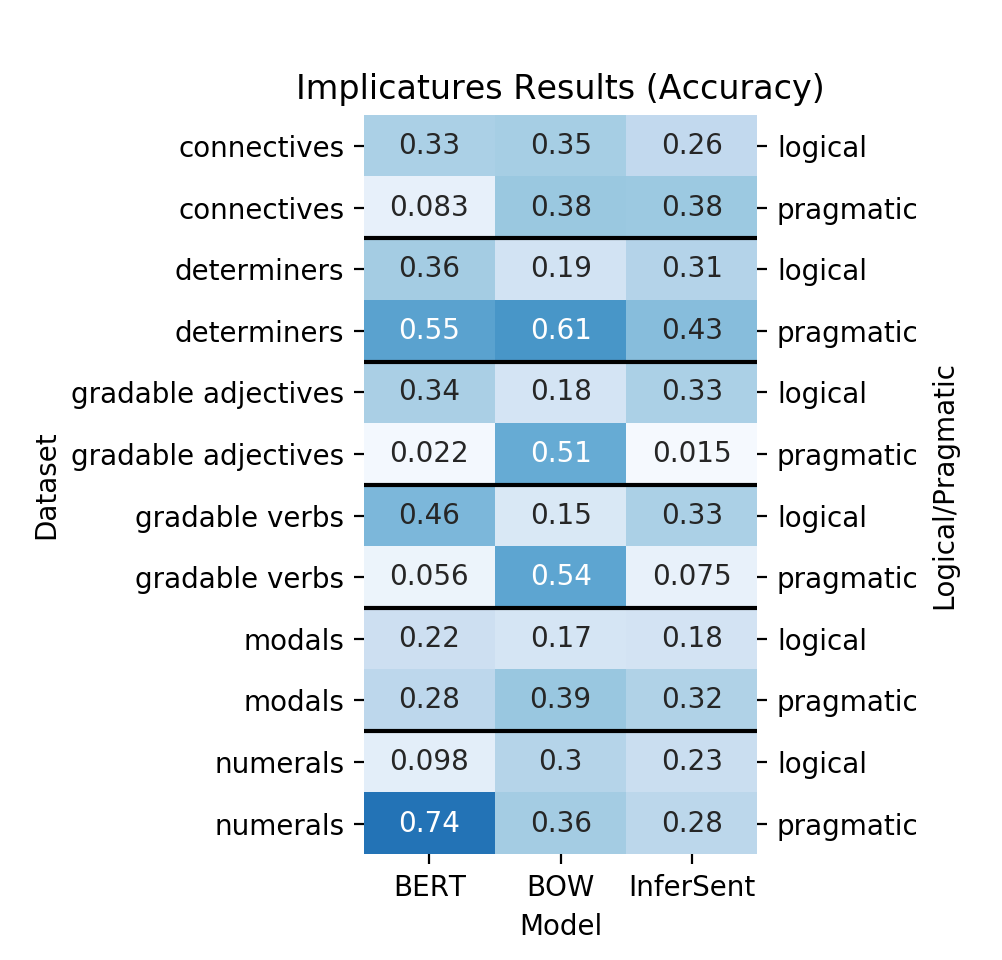}
    \caption{Results on Target Conditions (Implicatures)}\label{fig:si_results}
\end{figure}%

The scalar implicature portion of \DATASET includes six datasets%  we had 7!? but there are 6, someone else check this! <3 Adina
, each isolating a different scalar implicature trigger from six types of lexical scales (of the type described in \S \ref{sec:background}):
determiners $\langle${\it some, all}$\rangle$, connectives $\langle${\it or, and}$\rangle$, modals $\langle${\it can, have to}$\rangle$,  numerals $\langle$\textit{2,3}$\rangle$, $\langle$\textit{10,100}$\rangle$, scalar adjectives, and verbs, e.g.,\ $\langle${\it good, excellent}$\rangle$, $\langle${\it run, sprint}$\rangle$. Examples pairs of each implicature trigger can be found in Table~\ref{tab:appendix_SI} in the Appendix.
For each type, we generate 100 paradigms, each consisting of 12 unique sentence pairs, as shown in Table~\ref{tab:paradigm_SI}.

The six target sentence pairs comprise two main relation types: `implicature' and `negated implicature'. Pairs tagged as `implicature' have a premise that implicates the hypothesis (e.g.,\ \textit{some} and \textit{not all}). For `negated implicature', the premise implicates the negation of the hypothesis (e.g.,\ \textit{some} and \textit{all}), or vice versa (e.g.,\ \textit{all} and \textit{some}). 
% These conditions indirectly test implicature computation, and controlling for the models' knowledge of the asymmetric monotonicity between the items.
Six control pairs are logical contradictions, representing either scalar `opposites' (e.g., \textit{all} and \textit{none}), or `negations' (e.g., \textit{not all} and \textit{all}; \textit{some} and \textit{none}), probing the models' basic grasp of negation.

As mentioned in \S \ref{sec:implicature_background},  implicature computation is variable and dependent on the context of utterance. Thus, we anticipate two possible rational behaviors for a MultiNLI-trained model tested on an implicature: (a) be pragmatic, and compute the implicature, concluding that the premise and hypothesis are in an `entailment' relation, (b) be logical, i.e., consider only the literal content, and not compute the implicature, concluding they are in a `neutral' relation. Thus, we measure both possible conclusions, by tagging sentence pairs for scalar implicature with two sets of NLI labels to reflect the behavior expected under ``logical'' and ``pragmatic'' modes of inference, as shown in Table~\ref{tab:paradigm_SI}.  
%the ``logical'' label reflects valid inferences based on the semantic content of the sentences \emph{without} the computation of implicatures, while the ``pragmatic'' label includes any implicated content 

% \vspace{-1ex}

% \ex.\label{ex:pragmaticlogical}{{\it Premise}: Jo ate \textbf{some} of the cake.
% \\{\it Hyp.}: Jo did \textbf{not} eat \textbf{all} of the cake.
% \\Pragmatic label: {\sf\small entailment}
% \\Logical label: \sf\small neutral}

% \vspace{-1ex}

\subsection{Implicatures Results \& Discussion}\label{sec:results_SI}

%For most implicature datasets, results did not show interpretable pragmatic or logical behavior, with the exception of the determiner dataset, that we will devote more explanation for, after briefly commenting on the others (aggregate results per dataset in Figure \ref{fig:si_results}).% ; see Appendix for breakdown by condition).

We first evaluate model performance on the controls, shown in Figure~\ref{fig:si_controls}. Success on these controls is a necessary condition for us to conclude that a model has learned the basic function of negation (\emph{not}, \emph{none}, \emph{neither}) and the scalar relationship between terms like \emph{some} and \emph{all}. 
%, the other two models do less well, and have spotty coverage that it's hard to generalize about. 
We find that BERT performs at ceiling on control conditions for all implicature types, in contrast with InferSent and BOW, whose performance is very variable. 
Since only BERT passes all controls, its results on the target items are most interpretable. Full results for all models and target conditions by implicature trigger are in Figures~\ref{fig:adjectives}--\ref{fig:verbs} in the Appendix.

For connectives, scalar adjectives and verbs, the BERT model results correspond neither to the hypothesized pragmatic nor logical behavior.
In fact, for each of these subdatasets, the results are consistent with a treatment of scalemates (e.g.,\ \textit{and} and \textit{or}; \textit{good} and \textit{excellent}) as synonyms, e.g. it evaluates the `negated implicature' sentence pairs as `entailment' in both directions. This reveals a coarse-grained knowledge of these meanings that lacks information about asymmetric informativity relations between scalemates. Results for modals (\textit{can} and \textit{have to}) are split between the three labels, not showing any predicted logical or pragmatic pattern. We conclude that BERT has insufficient knowledge of the meaning of these words.

\begin{figure}[t]
    \centering
    \includegraphics[width=\columnwidth]{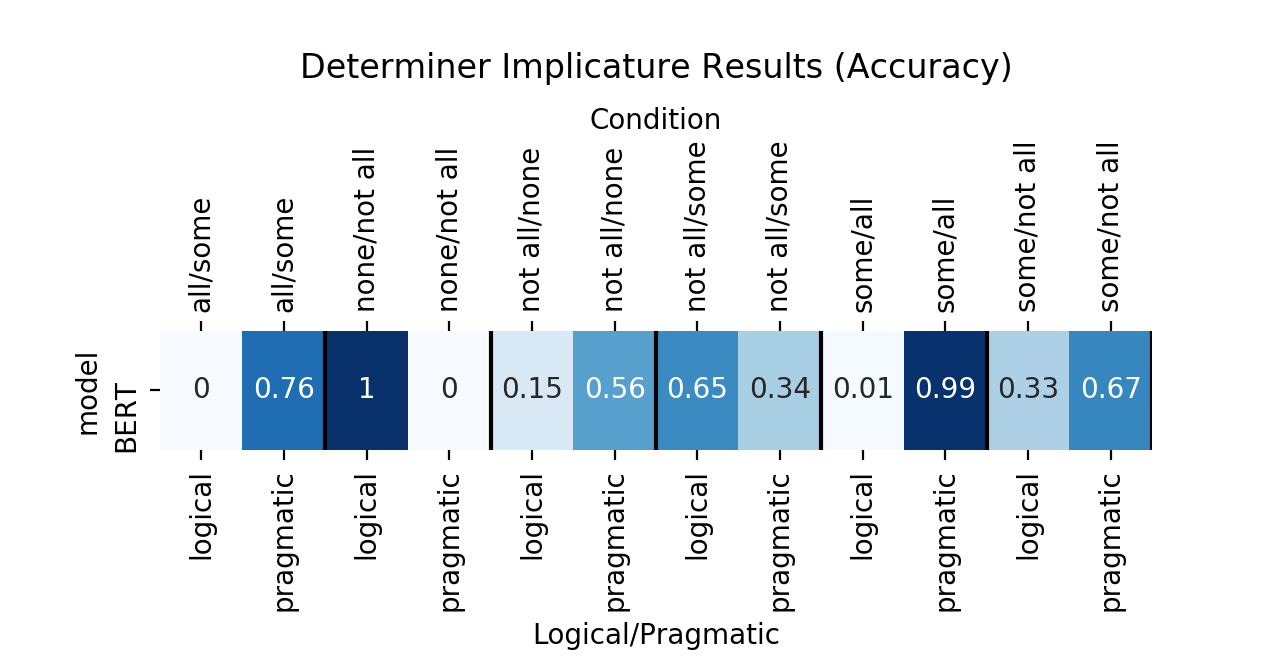}
    \caption{BERT results for scalar implicatures triggered by determiners $\langle${\it some, all}$\rangle$, by target condition.}
    \label{fig:determiner_results}
\end{figure}%

 In addition to pragmatic and logical interpretations, numerals can also be interpreted as exact cardinalities. We thus predict three different behaviors: logical ``at least $n$", pragmatic ``at least $n$", and  ``exactly $n$".
We observe that results are inconsistent: neither the ``exactly'' nor ``at least'' interpretations hold across the board.
% BERT predicts contradiction (at 100\%) of the time for \textit{2-3} and \textit{3-2}, in line with both exact and the pragmatic ``at least'' interpretation. However, it also predicts contradictions nearly 100\% of the time for \textit{not 2-not 3} and \textit{not 3-not 2}, which correspond only to the pragmatic ``at least'' interpretation. However, the control {\it 3-not 2} is rated only at 41\% contradiction, consistent with similar rates of ``exactly" and ``at least" readings.

For the determiner dataset (\textit{some}-\textit{all}), Figure~\ref{fig:determiner_results} breaks down the results by condition and shows that BERT behaves as though it performs pragmatic and logical reasoning in different conditions. Overall, it predicts a pragmatic relation more frequently (55\% vs.~36\%), and only 9\% of results are consistent with neither mode of reasoning. Furthermore, the proportion of pragmatic reasoning shows consistent effects of sentence order (i.e.,\ whether the implicature trigger is in the premise or the hypothesis), and the presence of negation in one or both sentences. %The breakdown of results per condition is in Figure~\ref{fig:determiner_results}.
Pragmatic reasoning is consistently higher when the implicature trigger is in the premise, which we can see in the results for negated implicatures: the \emph{some}--\emph{all} condition shows more pragmatic behavior compared to the \emph{all}--\emph{some} condition (a similar behavior is observed with the \emph{not all} vs. \emph{none} conditions).
% more li(e.g. \emph{all} and \emph{some}; \emph{some} and \emph{all}, where computing an implicature corresponds to a contradiction). 

% The order of weak and strong sentences has an effect on the proportion of logical vs. pragmatic reasoning:
% when the hypothesis is stronger than the premise (i.e. $P$: \textit{some} vs. $H$: \textit{all}; $P$: \textit{not all} vs. $H$: \textit{none}), pragmatic reasoning is higher for the same sentence in the opposite order. 

Generally, the presence of negation lowers rates of pragmatic reasoning. First, the negated implicature conditions can be subdivided into pairs with and without negation. Among the negated ones, pragmatic reasoning is lower than for non-negated ones.
Second, having negation in the premise rather than the hypothesis makes pragmatic reasoning lower: among pairs tagged as direct implicatures ({\it some} vs.\ {\it not all}), there is higher pragmatic reasoning with non-negated {\it some} in the premise than with negated {\it not all}.
Finally, we observe that pragmatic rates are lower for {\it some} vs. {\it not all} than for {\it some} vs. {\it all}. In this final case, pragmatic reasoning could be facilitated by explicit presentation of the two items on the scale.%\footnote{Interestingly, this parallels child behavior: \citet{ozturk-papafragou2015} show that children's difficulty in computing scalar implicatures with modals {\it may} and {\it have to} is eased when explicitly presented with the two alternatives.}
\begin{table}[t]
  \centering
\begin{adjustbox}{max width=\columnwidth}
{\small
\begin{tabular}{llllll}
\toprule
\multicolumn{2}{l}{\bf Presuppositions} & & \multicolumn{2}{c}{\bf Label} & {\bf Item} \\
Premise & Hypothesis & & & & {\bf Type}\\\midrule
    {*}Trigger & Prsp & & \multicolumn{2}{c}{entailment} & target\\
    {*}Trigger & Neg. Prsp & & \multicolumn{2}{c}{contradiction} & target\\
    {*}Trigger & Neut. Prsp & & \multicolumn{2}{c}{neutral} & target\\\midrule
    Neg. Trigger & Trigger & & \multicolumn{2}{c}{contradiction} & control\\
    Modal Trigger & Trigger & & \multicolumn{2}{c}{neutral} & control\\
    Interrog. Trigger & Trigger & & \multicolumn{2}{c}{neutral} & control\\
    Cond. Trigger & Trigger & & \multicolumn{2}{c}{neutral} & control\\
\bottomrule
\end{tabular}}
\end{adjustbox}
\caption{Paradigm for the presupposition target (top) and control datasets (bottom). For space, *Trigger refers to either plain, Negated, Modal, Interrogative, or Conditional Triggers as per Table~\ref{tab:prsp examples}.}\label{tab:paradigm}
\end{table}

%contains the stronger sentence (e.g., when the hypothesis contains \texit{all} and the premise contains \textit{some}), we see that the pragmatic interpretation is preferred.

In sum, for the datasets besides determiners, we find evidence that BERT fails to learn even the logical relations between scalemates, ruling out the possibility of computing scalar implicatures.  %were mostly treated as synonymous (except for modals, where knowledge of any semantic relationship between {\it can} and {\it need to} seemed absent).
It remains possible that BERT could learn these logical relations with explicit supervision \cite[see][]{richardson2020}, but it is clear that these are not learned from training on MultiNLI. Only the determiner dataset was informative in showing the extent of the NLI BERT model's pragmatic reasoning, since it alone showed a fine-grained enough understanding of the semantic relationship of the scalemates, like \textit{some} and \textit{all}.
% Within the determiner dataset, however, 
In this setting BERT returned impressive results showing a high proportion of pragmatic reasoning compared to logical reasoning, which was affected by sentence order and presence of negation in a predictable way.

\section{Experiment 2: Presuppositions}\label{sec:exp2}

\subsection{Presupposition Datasets}\label{subsec:presuppdata}
The presupposition portion of \DATASET includes eight datasets, each isolating a different kind of presupposition trigger. The full set of triggers is shown in Table~\ref{tab:appendix_prsp} in the Appendix. For each type, we generate 100 paradigms, with each paradigm consisting of 19 unique sentence pairs. %The paradigm structure is shown in Table~\ref{tab:paradigm}, and 
(Examples of the sentence types are in Table~\ref{tab:prsp examples}).

Of the 19 sentence pairs, 15 contain target items. The first target item tests whether the model correctly determines that the presupposition trigger entails its presupposition. The next two alter the presupposition, either negating it, or replacing a constituent, leading to contradiction and neutrality, respectively. The remaining 12 show that the relation between the trigger and the (altered) presupposition is not affected by embedding the trigger under various entailment-canceling operators. 4 control items are designed to test the basic effect of entailment-canceling operators---negation, modals, interrogatives, and conditionals. %In all cases, although $p$ entails $p$, $OP(p)$ does not entail $p$. 
In each control, the premise is a presupposition trigger embedded under an entailment-canceling operator, and the hypothesis is an unembedded sentence containing the trigger. These controls are necessary to establish whether models learn that presuppositions behave differently under these operators than do classical semantic entailments.

\subsection{Presupposition Results \& Discussion}\label{sec:results_prsp}

\begin{figure}[t]
    \centering
    \includegraphics[width=\columnwidth]{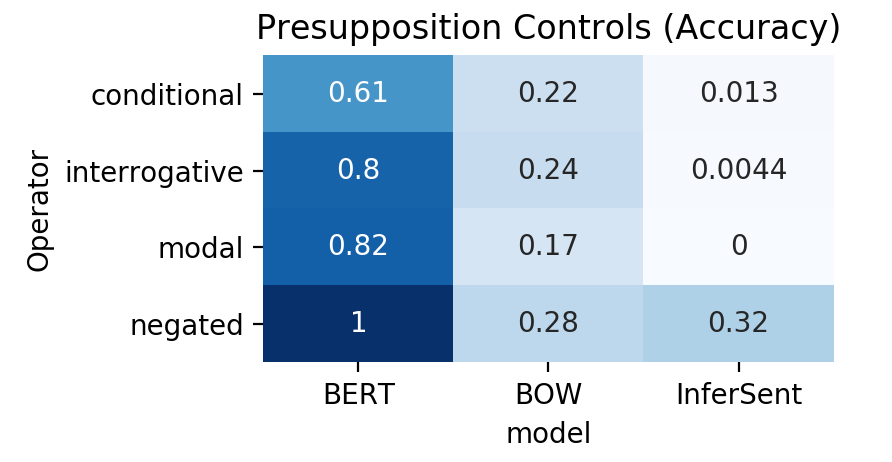}
    \caption{Results on Controls (Presuppositions). 
    % The premise contains a presupposition trigger embedded under the listed ``operator'', and the hypothesis contains an unembedded trigger.
    }%
    \label{fig:presupposition_controls}
\end{figure}%

The results from presupposition controls are in Figure~\ref{fig:presupposition_controls}. BERT performs well above chance on each control (acc.~$>0.33$), whereas BOW and InferSent perform at or below chance. In the ``negated'' condition, BERT correctly identifies that the trigger is contradicted by its negation 100\% of the time, e.g.,\ \emph{Jo's cat didn't go} contradicts \emph{Jo's cat went}. In the other conditions, it correctly identifies the neutral relation the majority of the time, e.g.,\ \emph{Did Jo's cat go?}~is neutral with respect to \emph{Jo's cat went}.\ This indicates that BERT mostly learns that negation, modals, interrogatives, and conditionals cancel classical entailments, while BOW and InferSent do not capture the ordinary behavior of these common operators.

\begin{figure}[t]
    \centering
    \includegraphics[width=\columnwidth]{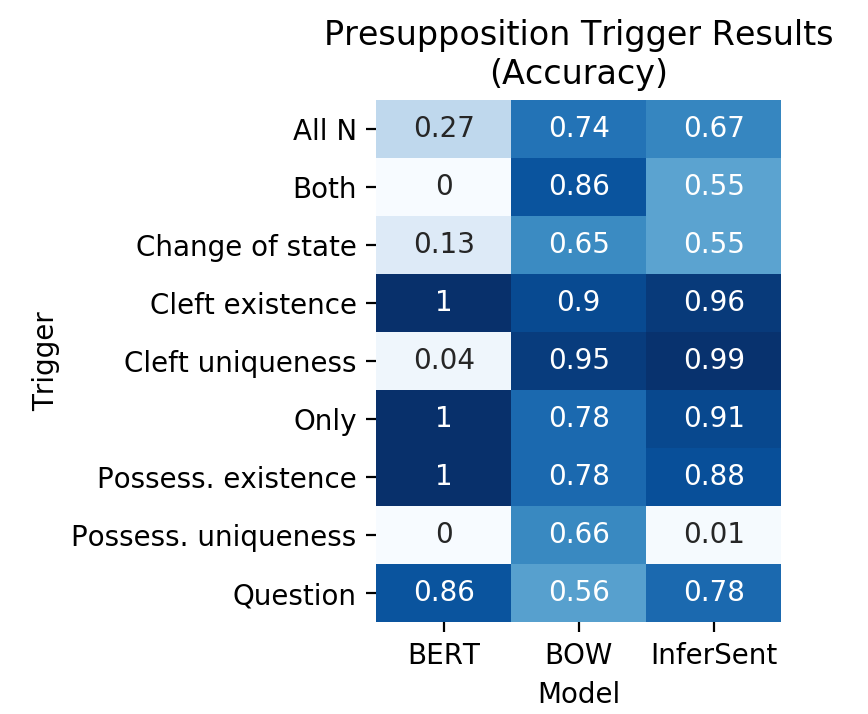}
    \caption{Results for the unembedded trigger paired with positive presupposition.}
    \label{fig:presupposition_trigger_results}
\end{figure}

Next, we test whether models identify presuppositions of the premise as entailments, e.g.,\ that \emph{Jo's cat went} entails that \emph{Jo has a cat}. 
Recall from \S \ref{sec:prsp} that this is akin to a listener accommodating a presupposition. The results in Figure~\ref{fig:presupposition_trigger_results} show that each of the three models accommodates some presuppositions, but this depends on both the nature of the presupposition and the model. For instance, the BOW and InferSent models accommodate presuppositions of nearly all trigger types at well above chance rates (acc.~\(\gg\) $33\%$). For the uniqueness presupposition of clefts, these models generally correctly predict an entailment (acc.~$>$ 90\%), but for most triggers, performance is less reliable. By contrast, BERT's behavior is bimodal. It always accommodates the existence presuppositions of clefts and possessed definites, as well as the presupposition of \emph{only}, but almost never accommodates any presupposition involving numeracy, e.g.\ \emph{Both flowers that bloomed died} entails \emph{There are exactly two flowers that bloomed}.\footnote{The presence of \emph{exactly} might contribute to poor performance on numeracy examples. We suspect MultiNLI annotators may have used it disproportionately for neut. hypotheses.} 

Finally, we evaluate whether models predict that presuppositions project out of entailment canceling operators (e.g., that \emph{Did Jo's cat go?}~entails that \emph{Jo has a cat}). We can only consider such a prediction as evidence of projection if two conditions hold: (a) the model correctly identifies that the relevant operator cancels entailments in the control from the same paradigm (e.g., \emph{Did Jo's cat go?}~is neutral with respect to \emph{Jo's cat went}), and (b) the model identifies the presupposition as an entailment when the trigger is unembedded in the same paradigm (e.g.\ \emph{Jo's cat went} entails \emph{Jo has a cat}). Otherwise, a model might correctly predict entailment essentially by accident if, for instance, it systematically ignores negation. For this reason, we filter out results for the target conditions that do not meet these criteria.

\begin{figure}
    \centering
    \includegraphics[width=\columnwidth]{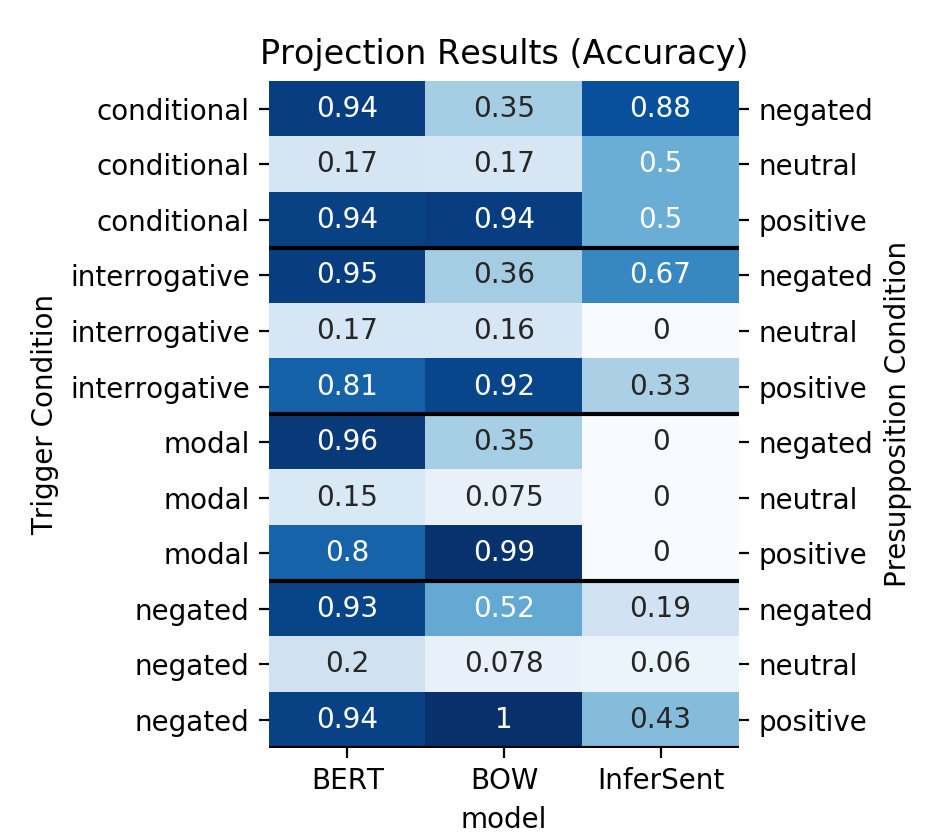}
    \caption{Results for presupposition target conditions involving projection.}
    \label{fig:presupposition_projection}
\end{figure}

 Figure~\ref{fig:presupposition_projection} shows results for the target conditions after filtering. While InferSent rarely predicts that presuppositions project, we find strong evidence that the BERT and BOW models do. Specifically, they correctly identify that the premise entails the presupposition (acc.~$\geq80\%$ for BERT, acc.~$\geq90\%$ for BOW). Furthermore, BERT is the only model to reliably identify (i.e., over 90\% of the time) that the negation of the presupposition is contradicted. These results hold irrespective of the entailment canceling operator. No model reliably performs above chance when the presupposition is altered to be neutral (e.g.,\ \emph{Did Jo's cat go?}~is neutral with respect to \emph{Jo has a cat}).

It is surprising that the simple BOW model can learn some of the projective behavior of presuppositions. One explanation for this finding is that many of the key features of presupposition projection are insensitive to word order. If a lexical presupposition trigger is present at all in a sentence, a presupposition will generally arise irrespective of its position in the sentence. There are some edge cases where this heuristic is insufficient, but \DATASET is not designed to test such cases.

To summarize, training on NLI is sufficient for all models we evaluate to learn to accommodate presuppositions of a wide variety of unembedded triggers, though BERT rejects presuppositions involving numeracy. Furthermore, BERT and even the BOW model appear to learn the characteristic projective behavior of some presuppositions.
% This is especially striking considering our findings in \S \ref{subsec:RelatedWork} that presuppositions are virtually absent from MultiNLI, indicating that this is learned through generalization. 

\section{General Discussion \& Conclusion}

We observe some encouraging results in \S \ref{sec:exp1}--\ref{sec:exp2}. 
We find strong evidence that BERT learns scalar implicatures associated with determiners {\it some} and {\it all}. Pragmatic or logical reasoning was not diagnosable for the other scales, whose meaning was not fully understood by our models (as most scalar pairs were treated as synonymous). 
In the case of presuppositions, the BERT NLI models, and BOW to some extent, perform well on a number of our subdatasets ({\it only}, cleft existence, possessive existence, questions). For the other subdatasets, the models did not perform as expected on the basic unembedded presupposition triggers, again suggesting the model's lack of knowledge of the basic meaning of these words. Though their behavior is far from systematic, this is suggestive evidence that some NLI models can  perform in ways that correlate with human-like pragmatic behavior.%, generalizing from data that does not have explicit examples of tested inference types. 

Given that MultiNLI contains few examples of the type found in \DATASET (see \S \ref{subsec:praginMultiNLI}), where might our positive results come from?
There are two potential sources of signal for the BERT model: NLI training, and pretraining (either BERT's masked language modeling objective or its input word embeddings). NLI training provides specific examples of valid (or invalid) inferences constituting an incomplete characterization of what commonsense inference is in general. Since presuppositions and scalar implicatures triggered by specific lexical items are largely absent from the MultiNLI data used for NLI training, any positive results on \DATASET would likely use prior knowledge from the pretraining stage to make an inductive leap that pragmatic inferences are valid commonsense inferences. The natural language text used for pretraining certainly contains pragmatic information, since, like any natural language data, it is produced with the assumption that readers are capable of pragmatic reasoning. Maybe this induces patterns in the data that make the nature of those assumptions recoverable from the data itself. %To the extent that pretraining underlies our results, we speculate that perhaps the shape of those assumptions is learnable from the text alone. 

This work is an initial step towards rigorously investigating the extent to which NLI models learn semantic versus pragmatic inference types. We have introduced a new dataset \DATASET for probing this question, which can be reused to evaluate pragmatic performance of any NLI given model.

%Overall, we conclude that, despite its original formulation as a semantic task, NLI models do learn some pragmatic reasoning. This suggests that the traditional dichotomy between semantics and pragmatics is not as evident to these models as we might have expected.

\section*{Acknowledgments} This material is based upon work supported by the National Science Foundation (NSF) under Grant No.~1850208 awarded to A. Warstadt. Any opinions, findings, and conclusions or recommendations expressed in this material are those of the author(s) and do not necessarily reflect the views of the NSF. Thanks to the FAIR NLP \& Conversational AI Group, the Google AI NLP group, and the NYU \href{https://wp.nyu.edu/ml2/}{ML$^{2}$}, including Sam Bowman, He He, Phu Mon Htut, Katharina Kann, Haokun Liu, Ethen Perez, Richard Pang, Clara Vania for discussions on the topic, and/or feedback on an earlier draft. Additional thanks to Marco Baroni, Hagen Blix, Emmanuel Chemla, Aaron Steven White, and Luke Zettlemoyer for insightful comments. % Adina says "when i talked at fair and google people gave advice. it doesn't hurt to thank, and it's sort of like a citation for your friends to get their names out. ofc one should only thank those who actually gave advice."

\bibliography{naaclhlt2018}
\bibliographystyle{acl_natbib}

\clearpage

% {\Large\textbf{Appendix}}
\section*{Appendix}

\begin{minipage}[t]{1.0\textwidth}

\begin{table}[H]
    \centering
    \begin{tabular}{lll}\toprule
    Type & Premise & Hypothesis \\ \midrule
        Connectives & These cats or those fish appear. & These cats and those fish don't both appear. \\
        Determiners & Some skateboards tipped over.&Not all skateboards tipped over.\\
        Numerals &Ten bananas were scorching.&One hundred bananas weren't scorching.\\
        Modals &Jerry could wake up.&Jerry didn't need to wake up.\\
        Scalar adjectives & Banks are fine. & Banks are not great.\\
        Scalar verbs & Dawn went towards the hills. & Dawn did not get to the hills.\\\bottomrule
        
    \end{tabular}
    \caption{The scalar implicature triggers in \DATASET. Examples are automatically generated sentences pairs from each of the six datasets for the scalar implicatures experiment. The pairs belong to the ``Implicature ($+$ to $-$)'' condition.}
    \label{tab:appendix_SI}
\end{table}

\begin{table}[H]
    \centering
    \resizebox{\textwidth}{!}{%
     {
    \begin{tabular}{lll}
    \toprule
    Type & Premise (Trigger) & Hypothesis (Presupposition)\\\midrule
       \emph{All N} & All six roses that bloomed died. & Exactly six roses bloomed. \\
       \emph{Both} & Both flowers that bloomed died. & Exactly two flowers bloomed.\\
       Change of State & The cat escaped. & The cat used to be captive.\\
       Cleft Existence & It is Sandra who disliked Veronica. & Someone disliked Veronica.\\
       Cleft Uniqueness & It is Sandra who disliked Veronica. & Exactly one person disliked Veronica.\\
       \emph{Only} & Only Lucille went to Spain. & Lucille went to Spain.\\
       Possessed Definites & Bill's handyman won. & Bill has a handyman.\\
       Question & Sue learned why Candice testified. & Candice testified.\\\bottomrule
    \end{tabular}
    }}
    \caption{The presupposition triggers in \DATASET. Examples are automatically generated sentences pairs from each of the eight datasets for the presupposition experiment. The pairs belong to the ``Plain Trigger / Presupposition'' condition.}
    \label{tab:appendix_prsp}
\end{table}

\end{minipage}

\newpage

\begin{figure*}[b]
    \centering
    \includegraphics[width=0.65\textwidth]{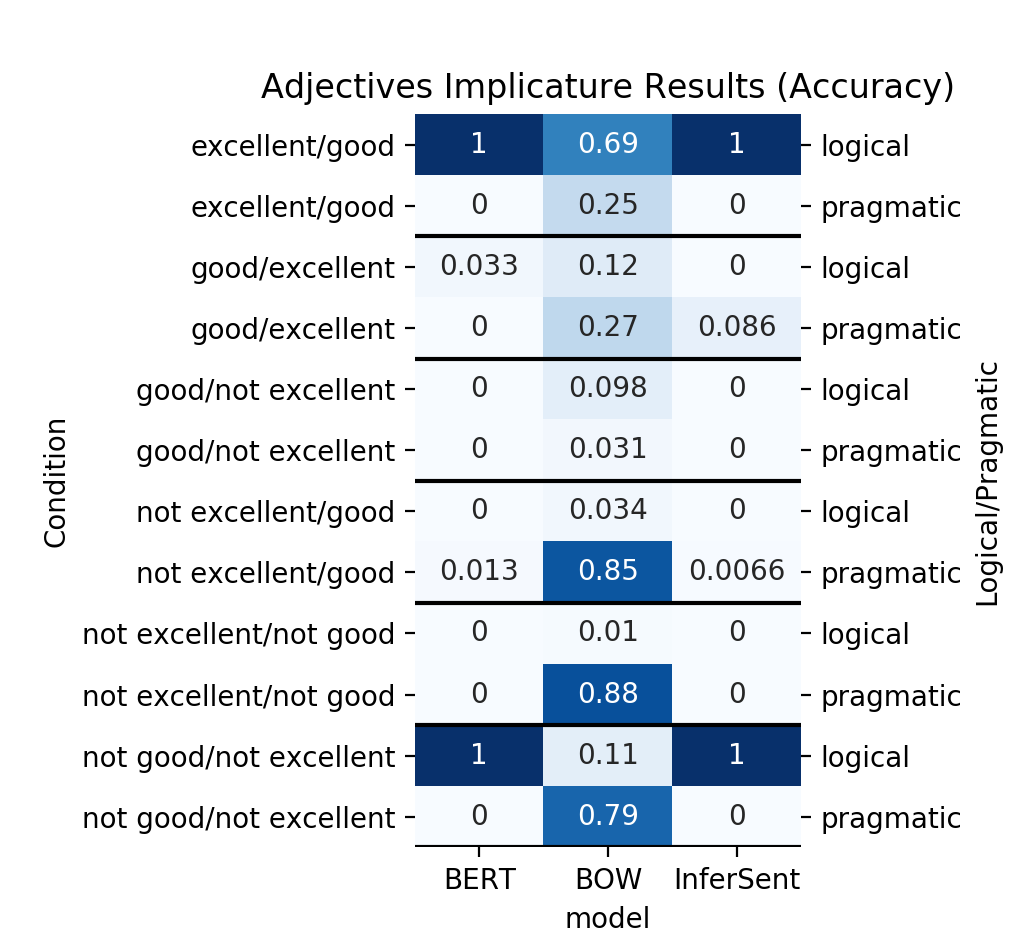}
    \caption{Results for the scalar implicatures triggered by adjectives, by target condition.}
    \label{fig:adjectives}
\end{figure*}

\begin{figure*}[h]
    \centering
    \includegraphics[width=0.65\textwidth]{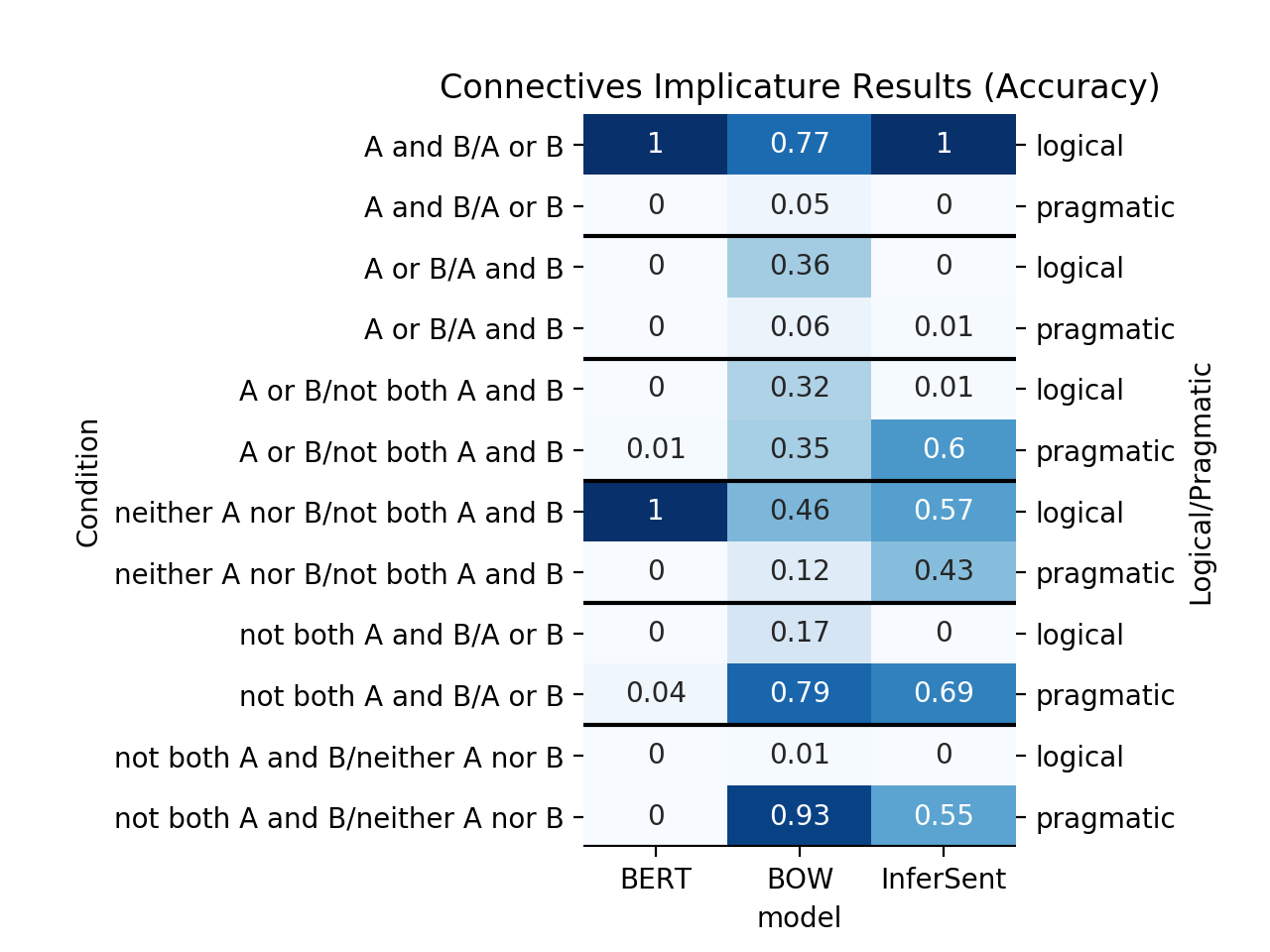}
    \caption{Results for the scalar implicatures triggered by adjectives, by target condition.}
    \label{fig:connectives}
\end{figure*}

\begin{figure}[h]
    \centering
    \includegraphics[width=\columnwidth]{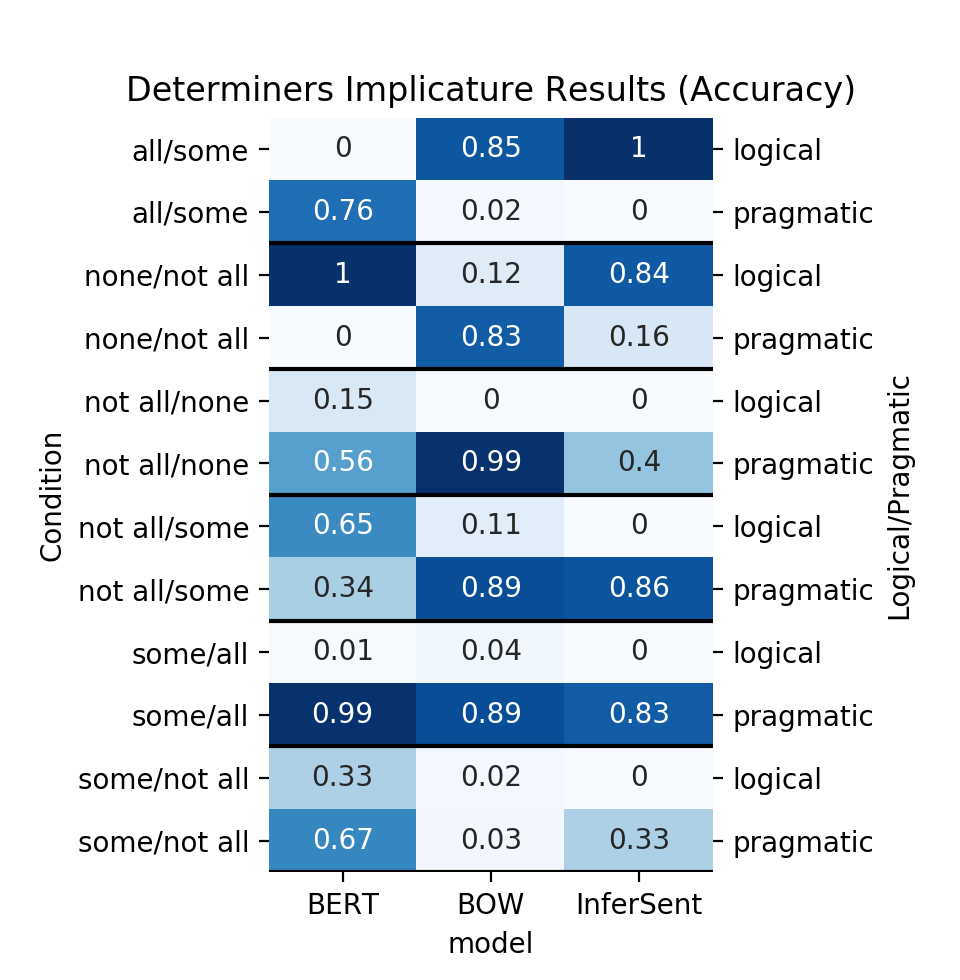}
    \caption{Results for the scalar implicatures triggered by determiners, by target condition.}
    \label{fig:determiner}
\end{figure}

\begin{figure}[h]
    \centering
    \includegraphics[width=\columnwidth]{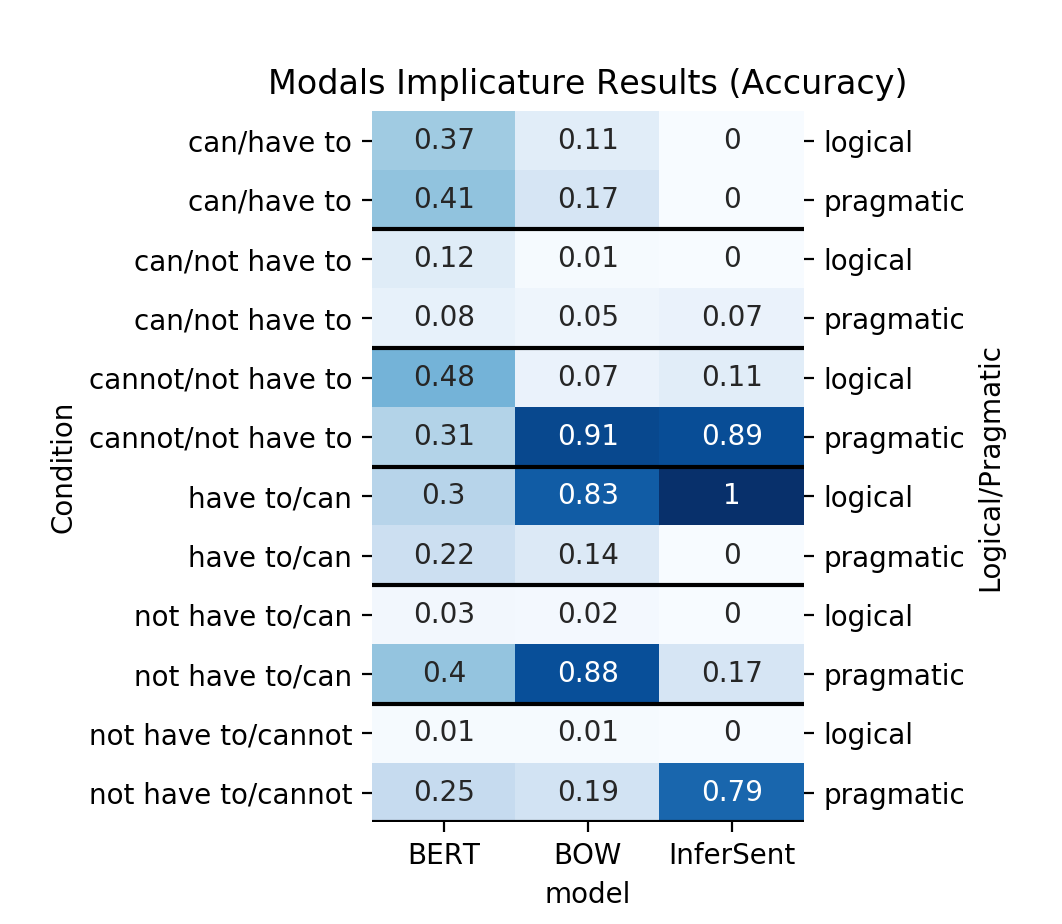}
    \caption{Results for the scalar implicatures triggered by modals, by target condition.}
    \label{fig:modals}
\end{figure}

\begin{figure}[h]
    \centering
    \includegraphics[width=\columnwidth]{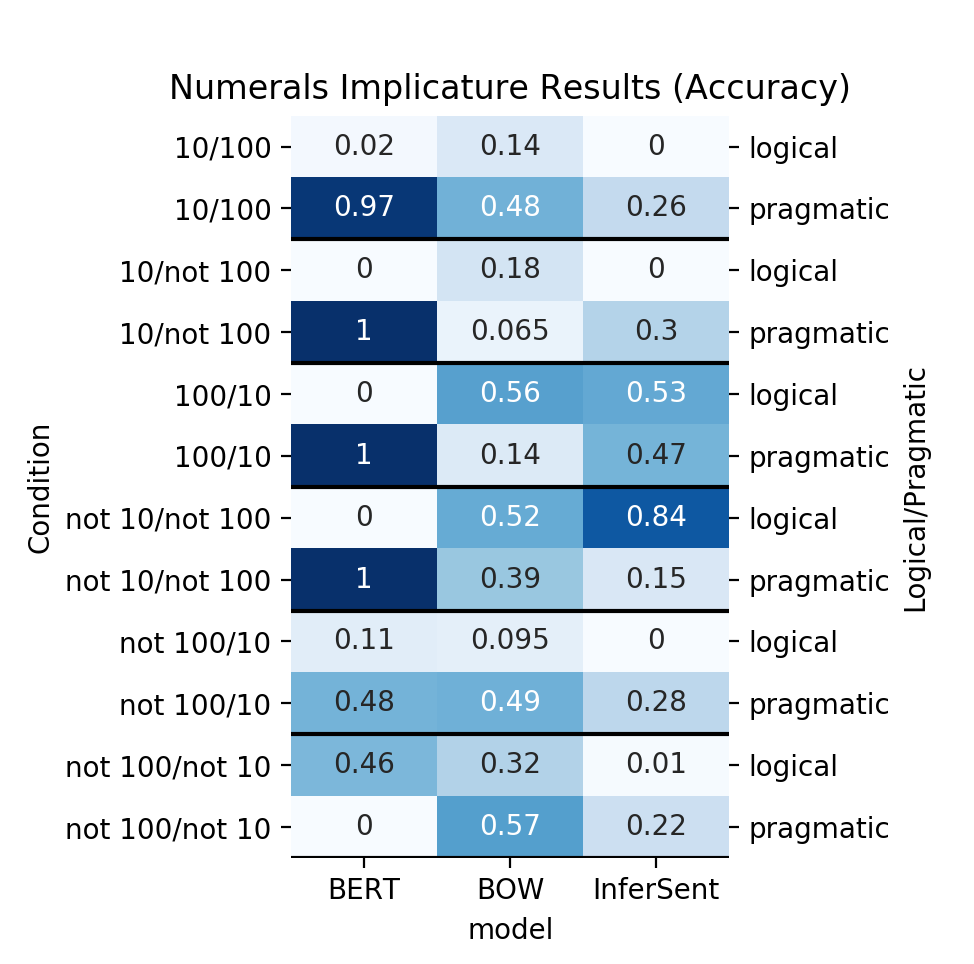}
    \caption{Results for the scalar triggered by numerals, by target condition.}
    \label{fig:numerals}
\end{figure}

\begin{figure}[h]
    \centering
    \includegraphics[width=\columnwidth]{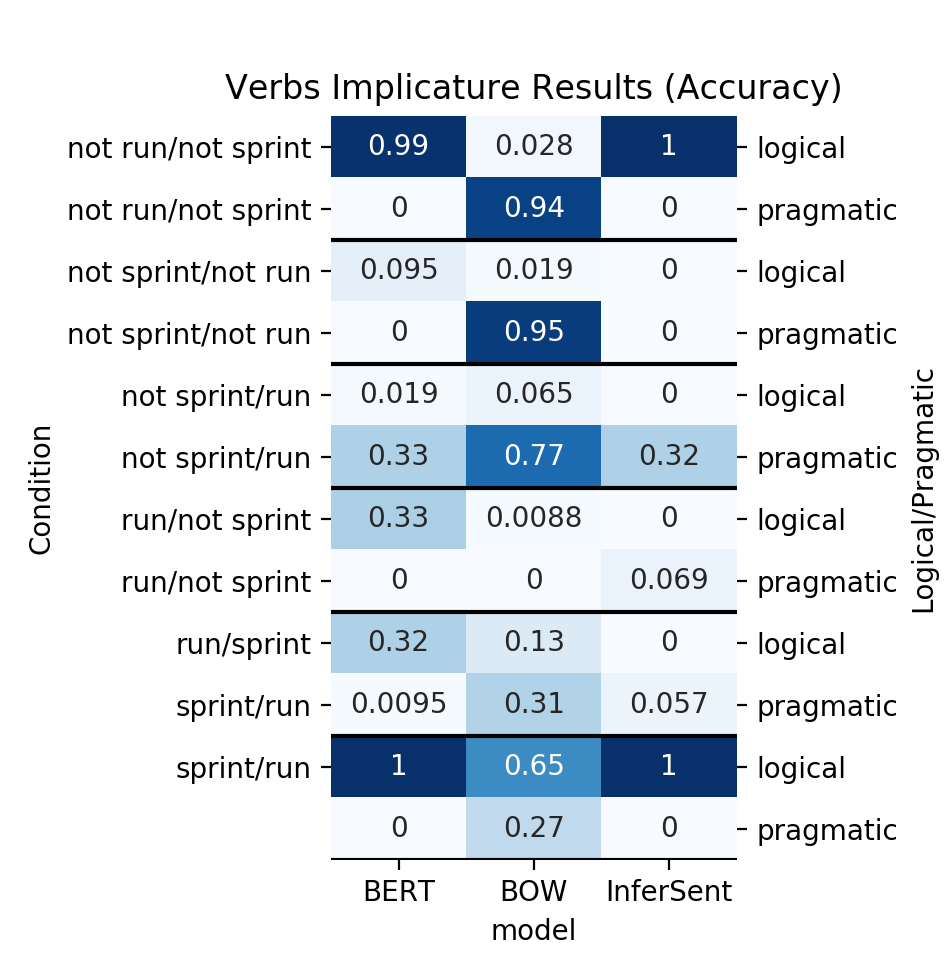}
    \caption{Results for the scalar implicatures triggered by verbs, by target condition.}
    \label{fig:verbs}
\end{figure}

\end{document}